\documentclass{article}
\usepackage{ijcai22}

\usepackage{times}

\usepackage{soul}
\usepackage{url}
\usepackage[hidelinks]{hyperref}
\usepackage[utf8]{inputenc}
\usepackage[small]{caption}
\usepackage{graphicx}
\graphicspath{{Figures/}}
\usepackage{bm}
\usepackage{amsmath}
\usepackage{amssymb}
\usepackage{booktabs}
\usepackage{threeparttable}
\usepackage{multicol}
\usepackage{multirow}

\title{Recent Advances and New Frontiers in Spiking Neural Networks}

\author{
 Duzhen Zhang\textsuperscript{\rm 1,2}\thanks{Corresponding Authors.}, Shuncheng Jia\textsuperscript{\rm 1,2}, Qingyu Wang\textsuperscript{\rm 1,2}
\affiliations
\textsuperscript{\rm 1}Institute of Automation, Chinese Academy of Sciences (CASIA), Beijing, China.\\
 \textsuperscript{\rm 2}School of Artificial Intelligence, University of Chinese Academy of Sciences
\emails
\{zhangduzhen2019\}@ia.ac.cn
}

\begin{document}

\maketitle

\begin{abstract}
In recent years, spiking neural networks (SNNs) have received extensive attention in brain-inspired intelligence due to their rich spatially-temporal dynamics, various encoding methods, and event-driven characteristics that naturally fit the neuromorphic hardware. With the development of SNNs, brain-inspired intelligence, an emerging research field inspired by brain science achievements and aiming at artificial general intelligence, is becoming hot. This paper reviews recent advances and discusses new frontiers in SNNs from five major research topics, including essential elements (i.e., spiking neuron models, encoding methods, and topology structures), neuromorphic datasets, optimization algorithms, software, and hardware frameworks. We hope our survey can help researchers understand SNNs better and inspire new works to advance this field.

\end{abstract}

\section{Introduction}
Brain science (BS) and artificial intelligence (AI) research have ushered in rapid development in mutual promotion, and brain-inspired intelligence research with interdisciplinary characteristics has received more and more attention. Brain-inspired intelligence aims to obtain inspiration from BS research regarding structure, mechanism, or function to improve AI. It enables AI to integrate various cognitive abilities and gradually approach or even surpass human intelligence in many aspects. Spiking neural networks (SNNs) are at the core of brain-inspired intelligence research. By emphasizing the highly brain-inspired structural basis and optimization algorithms, SNNs try to accelerate the understanding of the nature of biological intelligence from the perspective of computing, thereby laying a theoretical foundation for the formation of a new generation of human-level AI models.

As the main driving force of the current AI development, artificial neural networks (ANNs) have experienced multiple generations of evolution. The first generation of ANNs, called perceptron, can simulate human perception~\cite{rosenblatt1958perceptron}. The second generation is connectionism-based deep neural networks (DNNs) that emerged in the mid-1980s~\cite{rumelhart1986learning} and have led the development of AI for the past dozen years since 2006~\cite{hinton2006fast}. However, DNNs that transmit information primarily by firing rate are biologically imprecise and lack dynamic mechanisms within neurons. SNNs are considered the third generation of ANNs due to their rich spatially-temporal neural dynamics, diverse encoding methods, and event-driven advantages~\cite{maass1997networks}. 

The proposal of SNNs marks the gradual transition of ANNs from spatial encoding dominated by firing rate to spatially-temporal hybrid encoding dominated by precise spike firing and subthreshold dynamic membrane potential. The newly added temporal dimension makes it possible for more precise biological computing simulation, more stable and robust information representation, and more energy-efficient network computation.


 
In this survey, we comprehensively review the recent advances of SNNs and focus on five major research topics, which we define as:
\begin{itemize}
\item \textbf{Essential Elements.} The essential elements of SNNs include the neuron models as the basic processing unit, the encoding methods of the spike trains in neuron communication, and the topology structures of each basic layer at the network level (see Section~\ref{ee}).
\item \textbf{Neuromorphic Datasets.} The development of datasets has played an essential role in promoting the progress of ANNs. Currently, datasets suitable for SNNs are composed of spatially-temporal event streams, such as N-MNIST~\cite{orchard2015converting} and DVS-CIFAR10~\cite{li2017cifar10} (see Section~\ref{nd}).
\item \textbf{Optimization Algorithms.} How to efficiently optimize SNNs has been the focus of research in recent years. The research on optimization algorithms can be divided into two main types. One type is designed to understand better the biological system, such as spike-timing-dependent plasticity (STDP)~\cite{bi1998synaptic}. The other type is constructed to pursue superior computational performance, such as pseudo backpropagation (BP)~\cite{zenke2018superspike} (see Section~\ref{oa}).
\item \textbf{Software Frameworks.} Software frameworks are able to support the construction and training of SNNs, such as SpikingJelly~\cite{SpikingJelly} and CogSNN~\cite{RN846} (see Section~\ref{sf}).
\item \textbf{Hardware Frameworks.} Due to the advantage of ultra-low energy consumption of SNNs in hardware circuits, neuromorphic chips that support SNNs hardware implementation have sprung up, such as IBM TrueNorth~\cite{akopyan2015truenorth} and Intel Loihi~\cite{davies2018loihi} (see Section~\ref{hf}).
\end{itemize}

\begin{figure*}[tbp]
	\centering  
	\includegraphics[width=1.0\textwidth]{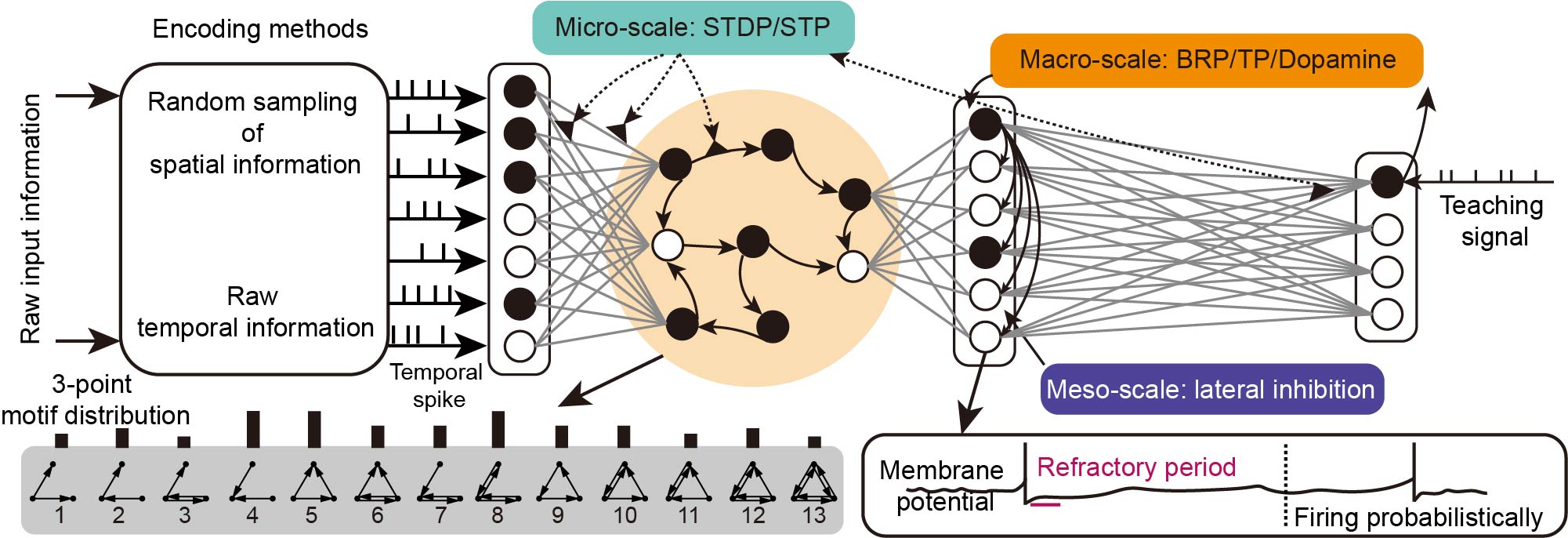}
	\caption{The overall architecture of SNNs, including encoding methods, motif topology, and multi-scale synaptic plasticity, etc.}
	\label{final_g}
\end{figure*}

We also have broad discussions on new frontiers in each topic. Finally, we summarize the paper (see Section~\ref{cc}). The overall architecture of SNNs is shown in Figure~\ref{final_g}. 

We hope this survey will help researchers understand the latest progress, challenges, and frontiers in the SNNs field.

\section{Essential Elements}\label{ee}
Essential elements include the neuron models as the basic information processing unit, the encoding methods of the spike trains in neuron communication, and the topology structures of each basic layer at the network level, which constitute the SNNs together.

\subsection{Neuron Models}
\subsubsection{Recent Advances} 

The typical structure of biological neurons mainly includes three parts: dendrite, soma, and axon~\cite{zhang2021neuron}. The function of dendrites is to collect input signals from other neurons and transmit them to the soma. The soma acts as a central processor, generating spikes when the afferent currents cause the neuron membrane potential to exceed a certain threshold (i.e., action potential). The spikes propagate along the axon without attenuation and transmit signals to the next neuron through the synapse at the axon's end. According to the dynamic characteristics of the neuronal potential, neurophysiologists have established many neuron models, represented by the Hodgkin-Huxley(H-H)~\cite{hodgkin1952quantitative}, the leaky integrate-and-fire (LIF)~\cite{dayan2003theoretical}, and the Izhikevich~\cite{izhikevich2004spike} model. 

By studying the potential data of squid axons, Hodgkin and Huxley~\shortcite{hodgkin1952quantitative} proposed a theoretical mathematical model of the mechanism of neuronal electrical activity, called the H-H model, formulated as:
\begin{equation}
\frac{dV}{dt} = -g_{N_a}(V-V_{N_a}) - g_K(V-V_K) + I\text{,}
\end{equation}
where $V$ denotes membrane potential, $g_{N_a}$ and $g_{K}$ denote the conductance densities of sodium and potassium ions, $V_{N_a}$ and $V_K$ denote the reversal potentials of sodium and potassium ion channels, and $I$ denotes total membrane current density. 

Since little is known about the mechanism of action potential generation in the early years, the process of action potential generation was simplified as follows: ``When the membrane potential exceeds the threshold $V_{th}$, the neuron will fire a spike, and the membrane potential falls back to the resting value $V_{rest}$". The LIF model follows this principle and introduces a leak factor, allowing the membrane potential to shrink over time~\cite{dayan2003theoretical}. It describes the change law of membrane potential below the threshold. The simplest and most common form of the LIF model is formulated as:
\begin{equation}
\tau_{m}\frac{dV}{dt} = V_{rest} - V + R_mI\text{,}
\end{equation}
where $\tau_m$ denotes membrane time constant, $V_{rest}$ denotes resting potential, and $I$ and $R_m$ denote the input current and the impedance of the cell membrane, respectively. The LIF model simplifies the action potential generation process but retains the three critical features of the biological neuron, i.e., membrane potential leakage, integration accumulation, and threshold firing. Later variants of the LIF model further describe the details of neuronal spiking activity, enhancing its biological credibility.

Izhikevich model uses a limited number of dimensionless parameter combinations (such as $a$ and $b$.) to characterize multiple types of rich spike firing patterns and can display the firing patterns of almost all known neurons in the cerebral cortex through the choice of parameters~\cite{izhikevich2004spike}. It is an efficient method for constructing second-order neural dynamics equations, formulated as:
\begin{align}
   \frac{dV}{dt} &= 0.04V^2 + 5V + 140 -u + I\text{,}\\
   \frac{du}{dt} &= a(bV - u)\text{,}
\end{align}
where $u$ is a membrane recovery variable used to describe the ionic current behavior, $a$ and $b$ are used to adjust the time scale of $u$ and the sensitivity to the membrane potential $V$.

By contrast, we briefly introduce the basic neuron model in DNNs. It retains the multi-input and single-output information processing form of biological neurons but further simplifies its threshold characteristics and action potential mechanism, formulated as:
\begin{equation}
y^l_i = \sigma(\sum_{j=0}^{n^{l-1}}w^{l-1}_{ij}x^{l-1}_j)\text{,}
\end{equation}
where after the weighted summation of the output values $x^{l-1}_j$ of the $n^{l-1}$ neurons in the previous layer, the nonlinear activation function $\sigma(\cdot)$ calculates the output value $y^l_i$ of the $i$th neuron in the $l$th layer. 

Compared with SNNs, DNNs use high-precision continuous floating-point values instead of discrete spike trains for communication, abandoning operations in the temporal domain and retaining only the spatial domain structure of the layer-by-layer computation. Although SNNs have lower expression precision, they keep richer neuron dynamics and are closer to biological neurons. In addition to receiving input in the spatial domain, the current state is also naturally influenced by past historical states. Therefore, SNNs have more substantial spatially-temporal information processing potential and biological plausibility. Due to threshold characteristics, the spike signal (0 or 1) of SNNs is usually very sparse, and the calculation is driven by events (only executed when the spike arrives), showing ultra-low power consumption and computational cost. Moreover, SNNs exhibit stronger anti-noise robustness than vulnerable DNNs. Individual neurons operating in spikes act as microscopic bottlenecks that maintain low intermittent noise and do not transmit sub-threshold noise to their neighbors. In summary, the spiking communication and dynamic characteristics of SNNs' neurons constitute the most fundamental difference from DNNs, which endows them with the potential for spatially-temporal task processing, ultra-low power, and robust computing. 

\subsubsection{New Frontiers} 

There are multiple choices of different abstraction levels between bionic degree and computational complexity for modeling biological neurons. Complex models, e.g., the H-H model, that use multi-variable, multi-group differential equations for precise activity descriptions cannot be applied to large-scale neural networks. Therefore, it is indispensable to simplify the model to speed up the simulation process. At present, the widely adopted LIF model can guarantee a low computational cost, but it relatively lacks biological credibility. To ensure the ability to construct larger-scale neural networks, it is still an urgent problem to find a neuron model with both excellent learning ability and high biological credibility.

\subsection{Encoding Methods}
\subsubsection{Recent Advances} 
At present, the common neural encoding methods mainly include rate, temporal, and population coding. Rate coding uses the firing rate of spike trains in a time window to encode information, where real input numbers are converted into spike trains with a frequency proportional to the input value~\cite{adrian1926impulses,cheng2020lisnn}. Temporal coding encodes information with the relative timing of individual spikes, where input values are usually converted into spike trains with the precise time, including time-to-first-spike coding~\cite{vanrullen2005spike}, rank order coding~\cite{thorpe1998rank}, etc. Besides that, population coding is special in integrating these two types. For example, each neuron in a population can generate spike trains with precise time and also contain a relation with other neurons for better information encoding at a global scale~\cite{georgopoulos1986neuronal,zhang2021population}.

\subsubsection{New Frontiers} 
Currently, the specific method of neural encoding has not yet been concluded. Populations of neurons with different encodings may coexist and cooperate, thus providing a sufficient perception of information. Neural encoding methods may behave differently in different brain regions. Compared to the limited case where current SNNs usually preset a single encoding method, more ideal and general SNNs should support hybrid applications of different encodings. They can utilize different encodings' advantages flexibly to optimize task performance, delay, and power consumption. Furthermore, many SNNs algorithms only pay attention to rate coding, ignoring the spike trains' temporal structure. It may cause that the advantages of SNNs in temporal information processing have not been well exploited. Therefore, the design of the algorithm suitable for temporal coding with high information density may be the new direction for future exploration.

\subsection{Topology Structures}
\subsubsection{Recent Advances} 

Similar to DNNs, the basic topology used to construct SNNs includes a fully connected, recurrent, and convolutional layer. The corresponding neural networks are multi-layer perceptrons (MLPs), recurrent neural networks (RNNs), and convolutional neural networks (CNNs). MLPs and RNNs are mainly for one-dimensional feature processing, while CNNs are mainly for two-dimensional feature processing. RNNs can be regarded as MLPs that add recurrent connections, and they are especially good at processing temporal features.

\subsubsection{New Frontiers} 

Compared to the structures in biological networks, the current topology of SNNs is relatively simplified. The structure of brain connections at different scales is very complex. Multi-point minimal motif network can be used as a primary network structure unit to analyze the functions of complex network systems~\cite{sporns2004motifs}. As Figure \ref{final_g} shows, taking the 3-point motif as an example, when the node types (such as different neuron types) are not considered, the combination of different primitive motifs is limited to 13 categories. For networks that complete similar functions, the Motif distributions tend to have strong consistency and stability. For function-specific networks, the Motif distributions between them are pretty different~\cite{sporns2004motifs}. Therefore, we can better understand their functions and connectivity patterns by analyzing the motif distributions in complex biological networks~\cite{jia2022motif}. Based on the parsed motif distributions, we can add constraints to the network structure design or search algorithms, thereby obtaining biologically plausible and interpretable new topology structures. Moreover, some recent methods propose an effective modeling framework to integrate SNNs with graph neural networks and exploit SNNs to process graph structure inputs~\cite{xu2021exploiting,zhu2022spiking}.

\section{Neuromorphic Datasets}\label{nd}
\subsection{Recent Advances} 

In the DNNs field, the continuous expansion of datasets in image, text, and other areas poses a challenge to the performance of DNNs but also promotes the development of DNNs from another perspective. The same is true for SNNs. The datasets inspired by neuromorphic vision sensors' imaging mechanisms are called neuromorphic datasets and are considered the most suitable dataset type for SNNs' applications.

Neuromorphic vision sensors (NVSs) inspired by biological visual processing mechanisms mainly capture light intensity changes in the visual field. They record spike train information in positive and negative directions according to the direction of information change, making NVSs have the characteristics of low latency, asynchronous, and sparse. Representative NVSs include dynamic vision sensors and dynamic, active imaging sensors. 

The following characteristics of neuromorphic datasets make them particularly suitable for benchmarking SNNs: 1) SNNs can naturally process asynchronous, event-driven information, making it a good fit with the data characteristics of neuromorphic datasets; 2) Temporal features embedded in neuromorphic datasets (such as precise firing times and temporal correlation between frames) provide an excellent platform to demonstrate the ability to spiking neurons to process information via spatially-temporal dynamics. 

According to the dataset construction method, the current neuromorphic datasets are mainly divided into three categories. The first category is the datasets collected from the field scene, which are primarily captured directly by NVSs to generate unlabeled data, such as DvsGesture~\cite{amir2017low} for gesture recognition. The second category is the transformation datasets, mainly generated from the labeled static image datasets through the actual shooting of NVSs,  such as DVS-CIFAR10~\cite{li2017cifar10}. Due to their ease of use and evaluation, such transformation datasets are the most commonly used datasets in SNNs. The third category is the generated datasets, which are mainly generated using labeled data through algorithms that simulate the characteristics of NVSs. They generate neuromorphic datasets directly from the existing image or video stream information through a specific difference algorithm~\cite{bi2017pix2nvs}.
 
\subsection{New Frontiers} 

Although the research on neuromorphic datasets is still developing, these three categories of datasets have their limitations. Due to inconsistencies in the way researchers preprocessed the first category datasets, such as time resolution and image compression scale, the results reported so far are difficult to compare fairly. The second and third categories of datasets are mainly generated by the secondary transformation of the original static data, and their data is difficult to express rich temporal information. Therefore, they cannot fully use the spatially-temporal processing characteristics of SNNs. In summary, the current research on neuromorphic datasets is still in its infancy. 

On the static image datasets in the DNNs field, e.g., MNIST~\cite{lecun1998mnist}, CIFAR~\cite{krizhevsky2009learning}, etc., the performance of SNNs is usually not as good as that of DNNs. However, studies have shown that it is unwise to blindly measure SNNs on such datasets with a single criterion, e.g., classification accuracy~\cite{deng2020rethinking}. In datasets that contain more dynamic temporal information and naturally have the form of spike signals, SNNs can achieve better results in terms of performance and computational overhead. As mentioned above, the small-scale datasets obtained by NVSs are the mainstream of current SNN datasets, but it does not rule out that there may be other more suitable data sources to be explored. In addition to the simple image recognition task, it is hoped to develop spatially-temporal event flow datasets ideal for diverse tasks to investigate further the potential advantages and possible application scenarios of SNNs. Moreover, constructing larger-scale and more functionally fit datasets (fully exploiting the spatially-temporal processing capabilities of spiking neurons and the event-driven properties of data) is also an important future direction to provide a broad and fair benchmark for SNNs.

\section{Optimization Algorithms}\label{oa}
\subsection{Recent Advances}
The learning of ANNs is to optimize network parameters based on task-specific datasets. Optimization algorithms play a crucial role in it. In DNNs, gradient-based error BP optimization algorithms~\cite{rumelhart1986learning} are the core of the current DNNs optimization theory and are widely used in practical scenarios. In contrast, there is no recognized core optimization algorithm in the SNNs field. There are different emphases between biological plausibility and task performance. In addition, different neuron models, encoding methods, and topological structures used in the network all lead to the diversification of optimization algorithms. The research on optimization algorithms of SNNs can be divided into two main types. One type is designed to understand better the biological system, where detailed biologically-realistic neural models are used without further consideration of computational performance. The other type is constructed to pursue superior computational performance, where only limited features of SNNs are retained, and some efficient but not biologically-plausible tuning algorithms are still used. 


The first category of algorithms satisfies known BS discoveries as much as possible. This paper innovatively further divides them into plasticity optimization based on micro-scale, meso-scale and macro-scale. Micro- and meso-scale plasticity are typically self-organizing, unsupervised local algorithms, and macro-scale plasticity is typically supervised global algorithms. Micro-scale plasticity mainly describes the properties of learning that take place at a single neuron or synaptic site, including STDP~\cite{bi1998synaptic}, short term plasticity (STP)~\cite{zhang2018brain}, Reward-STDP~\cite{mozafari2019spyketorch}, Dale rule~\cite{yang2016dendritic}, etc. Such algorithms achieve decent performance on simple image classification tasks. Diehl~\shortcite{diehl2015unsupervised} uses two-layer SNNs with LIF neurons, and the adjacent layer neurons use STDP for learning, achieving a test accuracy of 95\% on MNIST. It is subsequently optimized to 96.7\% accuracy by incorporating plasticity mechanisms such as symmetric-STDP and dopamine modulation~\cite{hao2020biologically}. Kheradpisheh~\shortcite{kheradpisheh2018stdp} uses multi-layer convolution, STDP, and information delay for efficient image feature classification, achieving 98.40\% accuracy on MNIST. Moreover, a combined optimization algorithm of STDP and Reward-STDP is proposed to optimize multi-layer spiking convolutional networks~\cite{mozafari2019spyketorch}. Meso-scale plasticity mainly describes the relationship between multiple synapses and multiple neurons, e.g., lateral inhibition~\cite{zenke2015diverse}, Self-backpropagation (SBP)~\cite{zhang2021self}, homeostatic control among multiple neurons, etc. Zhang~\shortcite{zhang2018plasticity} proposes an optimization algorithm based on neural homeostasis to stabilize a single node's input and output information. Macro-scale plasticity mainly describes the top-down global credit distribution. Unfortunately, there is no global optimization algorithm similar to BP in the credit assignment of biological networks. The directionality of synaptic information transmission makes forward transmission and possible feedback pathways physiologically separate. The brain has no known way to access forward weights during backpropagation, which is called the weight transport problem. To make BP more biological-like and energy-efficient, some transformative algorithms for BP have emerged. For example, target propagation~\cite{bengio2014auto}, feedback alignment~\cite{lillicrap2016random}, direct random target propagation~\cite{frenkel2021learning}, etc., solve the weight transport problem by implementing direct gradient transfer in the backward process with random matrices. They bring new ideas to the plasticity optimization of SNNs at the macro-scale, e.g., biologically-plausible reward propagation (BRP)~\cite{zhang2021tuning}.

The second category of algorithms typically employs different BP-based variants for the optimization of SNNs, mainly including pseudo-BP~\cite{zenke2018superspike}, DNNs-converted SNNs~\cite{cao2015spiking}, etc. Since the spike signal is not differentiable, the direct application of gradient-based BP is difficult. The key feature of the pseudo-BP is replacing the non-differential parts of spiking neurons during BP with a predefined gradient number~\cite{shrestha2018slayer}. On some smaller-scale datasets, its performance and convergence speed are comparable to DNNs after standard BP training. The basic idea of DNNs-converted SNNs is that the average firing rate under rate encoding in SNNs can approximate the continuous activation value under the ReLU activation function in DNNs. After the original DNNs are trained with BP, it is converted into SNNs by specific means~\cite{cao2015spiking,deng2020optimal,li2022efficient,li2022spike}. In terms of performance, the DNNs-converted SNNs maintain the smallest gap with DNNs and can be implemented on large-scale network structures and datasets. For example, Rueckauer~\cite{rueckauer2017conversion} implements some spiking versions of the VGG-16 and GoogLeNet models. Sengupta~\cite{sengupta2019going} reports that the VGG-16 achieved 69.96\% accuracy on the ImageNet dataset with a conversion precision loss of 0.56\%. Subsequently, Hu~\cite{hu2018spiking} uses the deep structure of ResNet-50 to obtain 72.75\% accuracy.


\subsection{New Frontiers} The organic combination of biological plausibility and performance will remain the relentless goal of SNNs optimization algorithms. Compared with DNNs, only a few algorithms can directly train truly large-scale deep SNNs. Problems such as gradient vanishing, high resource overhead, and even non-convergence in deep networks training need further exploration. Recently, residual learning and batch normalization from the DNNs field has been introduced into pseudo-BP to train deep SNNs directly~\cite{fang2021deep,hu2021advancing,zheng2021going}, which achieves excellent results and may serve as a way for the optimization development of deep SNNs in the future. Moreover, existing DNNs-converted SNNs algorithms also suffer from long simulation periods. From the perspective of model compression, the conversion process is an extreme quantization of activation values. The binary neural networks (BNNs)~\cite{rastegari2016xnor,DBLP:conf/ijcnn/ChenLSXX18} in DNNs have a similar concept. However, the connection and difference between the BNNs and SNNs, and the possible impact of the additional temporal dimension in SNNs are not clearly elaborated. The threshold firing properties of SNNs may make them more receptive to compression algorithms. Therefore, the combination with compression algorithms such as weight quantization and pruning also needs to be explored so that the computational efficiency advantage of SNNs can be further developed~\cite{chen2021pruning}.

\section{Software Frameworks}\label{sf}
\subsection{Recent Advances} 

The software framework of SNNs is a programming tool used to help the SNNs to achieve rapid simulation, network modeling, and algorithm training. Software frameworks are related to the effective reduction of the field entry threshold and the efficient development of large-scale SNNs projects, providing substantial support for the scientific research of SNNs. Due to the differences in research goals and implementation methods, there are many software frameworks.  

Some software frameworks have been used to achieve smaller-scale neuronal functional simulations, mainly to understand biological systems. Neuron~\cite{migliore2006parallel} and Nest~\cite{gewaltig2007nest} are two commonly used software frameworks that support multiple programming languages, e.g., Python, C++, etc., and visual interfaces. They also support the characterization of more detailed neuronal activity dynamics, such as H-H, LIF, and Izhikevich, or multi-compartmental models with complex structures. Other frameworks can implement task-specific, larger-scale SNNs optimization calculations, support multiple neuron models, and support various types of synaptic plasticity, such as STDP and STP. For example, Bindsnet~\cite{hazan2018bindsnet}, Brain2~\cite{stimberg2019brian}, Spyketorch~\cite{mozafari2019spyketorch}, SpikingJelly~\cite{SpikingJelly}, CogSNN~\cite{RN846}, etc., based on the Python language, better support multi-neuron networking, and can be used for some relative complex pattern recognition tasks. In particular, CogSNN~\cite{RN846} has excellent support capabilities for neuromorphic datasets introduced earlier, such as N-MNIST, DVS-CIFAR10, and DvsGesture, and supports cognitive computations such as the Muller-Lyer illusion and McGurk effect. 

\subsection{New Frontiers} 

The software frameworks of SNNs are still in a relatively primary development stage. In DNNs, many kinds of software frameworks support their training, and the common one is PyTorch~\cite{paszke2019pytorch}. The user-friendly programming interface and the unified data stream processing method make it easy for a beginner to build and train DNNs, which significantly promotes the development of the DNNs field. However, in SNNs, currently, only a few frameworks can support the construction and training of large-scale SNNs. Building large SNNs still requires programmers to have excellent programming skills. Therefore, the development of user-friendly programming frameworks to effectively deploy large-scale SNNs is crucial to the development of this field.

\section{Hardware Frameworks}\label{hf} 
\subsection{Recent Advances} 

The development of the SNNs software frameworks enables the corresponding applications to be extended to more and more practical scenarios quickly. In particular, application scenarios with high demands for limited size, low energy consumption, parallel computing, etc., such as robot chips, high-performance analog computing, pattern recognition accelerators, and event high-speed cameras, have gradually begun to show great application potential. 

Since SNNs have the advantage of ultra-low energy consumption in hardware circuits, in the last ten years, neuromorphic chips that support SNNs hardware implementation, represented by TrueNorth~\cite{akopyan2015truenorth}, Loihi~\cite{davies2018loihi}, and Tianjic~\cite{pei2019towards} chips, have sprung up. Unlike the traditional Von Neumann processor architecture, many computing cores work simultaneously in a neuromorphic chip, exchanging intermediate results through a routing network. The whole system usually does not have a unified external memory. Instead, each computing core has its own independent storage space, presenting a decentralized operation mode, so it has incredibly high parallelism and memory access efficiency.

The existing neuromorphic chips can be divided into offline and online chips according to whether they support learning functions. Offline chips mean that the parameters (e.g., weights) of SNNs have been trained in advance. The model only needs to be deployed to the neuromorphic chips, and the parameters will not be updated in the subsequent running process. That said, the offline chips only support the inference process of SNNs, but not their training process. Such chips include TrueNorth~\cite{akopyan2015truenorth}, Tianjic~\cite{pei2019towards}, and Neurogrid~\cite{benjamin2014neurogrid}. Unlike the offline chips, the online chips support parameter updates during the running process of the SNNs model. Such chips include Loihi~\cite{davies2018loihi}, SpiNNaker chip~\cite{furber2014spinnaker}, and some developing chips equipped with CogSNN toolbox~\cite{RN846}.


IBM TrueNorth chip~\cite{akopyan2015truenorth} contains about 5.4 billion silicon transistors, 4096 cores, 1 million neurons, and 256 million synapses, which can realize applications such as SNNs-based brain-inspired affective computing. The number of neurons in the Intel Loihi chip has reached 8 million, and synapses have reached 8 billion~\cite{davies2018loihi}. It initially plays a role in highly sensitive odor perception and recognition. Stanford University Neurogrid chip simulates millions of neurons connected by billions of synapses in real-time, supporting high-performance computers and brain-like robot chips~\cite{benjamin2014neurogrid}. Its current progress includes high-throughput brain information processing, brain-computer interface neural information recording, etc. The DNNs/SNNs hybrid brain-inspired Tianjic chip developed by Tsinghua University can support traditional DNNs, and a new generation of SNNs~\cite{pei2019towards}. They also verified the functions of speech recognition, control tracking, and automatic obstacle avoidance of Tianjic on self-driving bicycles. 

\subsection{New Frontiers} 

In the context of the Von Neumann bottleneck, as an alternative computing paradigm to traditional digital circuits, neuromorphic chips have become a research hotspot for more than ten years and have achieved fruitful results. By drawing inspiration from the brain's structure and function, neuromorphic chips provide an efficient solution for event-driven computation in SNNs, achieving essential properties such as high parallelism and ultra-low power consumption. Combining various advantageous technologies of existing hardware chips is an important direction that requires in-depth study. This cross-integration may be reflected in the following aspects: 1) Heterogeneous fusion of two paradigms, DNNs and SNNs, improves overall performance; 2) Mixed-precision computation of low-precision memristors and high-precision digital circuits; 3) A general-purpose computing chip combines high-efficiency but low-performance unsupervised local learning with high-performance but low-efficiency supervised global learning.

\section{Conclusion}\label{cc}
In this paper, we provide a literature survey for SNNs. We review the recent advances and discuss the new frontiers in SNNs from five major research topics: essential elements (i.e., neuron models, encoding methods, and topology structures), neuromorphic datasets, optimization algorithms, software, and hardware frameworks. We hope that this survey can shed light on future research in the SNNs field.

\section*{Acknowledgments}
This work was supported by the National Key R\&D Program of China (2020AAA0104305), the Shanghai Municipal Science and Technology Major Project, and the Strategic Priority Research Program of the Chinese Academy of Sciences (XDA27010404, XDB32070000).


\small{
\bibliographystyle{named}
\bibliography{ijcai22}
}

\end{document}